\documentclass[10pt,twocolumn,letterpaper]{article}

\usepackage[pagenumbers]{cvpr} 

\definecolor{cvprblue}{rgb}{0.21,0.49,0.74}
\usepackage[pagebackref,breaklinks,colorlinks,allcolors=cvprblue]{hyperref}

\def\paperID{13530} 
\def\confName{CVPR}
\def\confYear{2026}

\title{DisDop: Distillation with Domain Priors for Open-Vocabulary \\ Aerial Object Detection}

\author{
Ruihao Xu\textsuperscript{*1}\quad
Yong Liu\textsuperscript{*1}\quad
Yansong Tang\textsuperscript{1}\quad
Sule Bai\textsuperscript{1}\quad
Xubing Ye\textsuperscript{1}\\
Bingyao Yu\textsuperscript{1}\quad
Yutao Guo\textsuperscript{1}\quad
Jiwen Lu\textsuperscript{2}\quad
Jie Zhou\textsuperscript{\dag\,2}\\
\textsuperscript{1}Tsinghua Shenzhen International Graduate School, Tsinghua University\\
\textsuperscript{2}Tsinghua University
}

\usepackage{xcolor}         
\usepackage{stfloats}
\usepackage{amssymb} 
\usepackage{pifont} 

\usepackage{makecell}
\usepackage{multirow}
\usepackage{graphicx}
\usepackage{colortbl}
\usepackage{amsmath}
\usepackage{caption}
\usepackage{booktabs}

\usepackage{algorithm}
\usepackage{algpseudocode}
\algtext*{EndIf}
\algtext*{EndFor}

\begin{document}
\maketitle
\let\thefootnote\relax
\footnotetext{\textsuperscript{*}Equal contribution. \textsuperscript{\dag}Corresponding author.}
\begin{abstract}
With the widespread application of drones in recent years, object detection of aerial images has attracted increasing attention, especially open-vocabulary aerial detection which is not restricted to predefined categories. 
Due to the scarcity of drone's viewpoint images and their significant differences from natural images, it is difficult to achieve satisfying results by directly applying vanilla open-vocabulary detection methods designed for natural scenarios. 
Some studies propose to transfer knowledge from pre-trained models by using lightweight networks or generating pseudo labels, but they tend to rely on models trained on natural images, neglecting the potential of foundation models specifically tailored for remote sensing and aerial imagery.
To address this limitation, we propose DisDop, a unified framework that systematically distills multi-level domain priors from remote sensing foundation models into a lightweight detector.
Specifically, we first distill visual priors through a teacher fusion strategy that combines RemoteCLIP's cross-modal alignment capability with DINOv3's fine-grained local feature extraction ability, transferring their complementary strengths to the detector's backbone.
Second, we distill textual priors embedded in RemoteCLIP's text encoder by explicitly modeling inter-category semantic relationships, while incorporating global contextual priors to enhance local feature representation for small objects.
Through this multi-level prior distillation framework, our DisDop achieves new state-of-the-art performance on open-vocabulary aerial detection benchmarks.
Extensive ablation analysis also demonstrates the effectiveness of our proposed modules.
Code is available at \href{https://github.com/Rio-Allen/DisDop}{here}.
\end{abstract}
\section{Introduction}
\label{sec:intro}

\begin{figure}[!t]
\centering
\includegraphics[width=\linewidth]{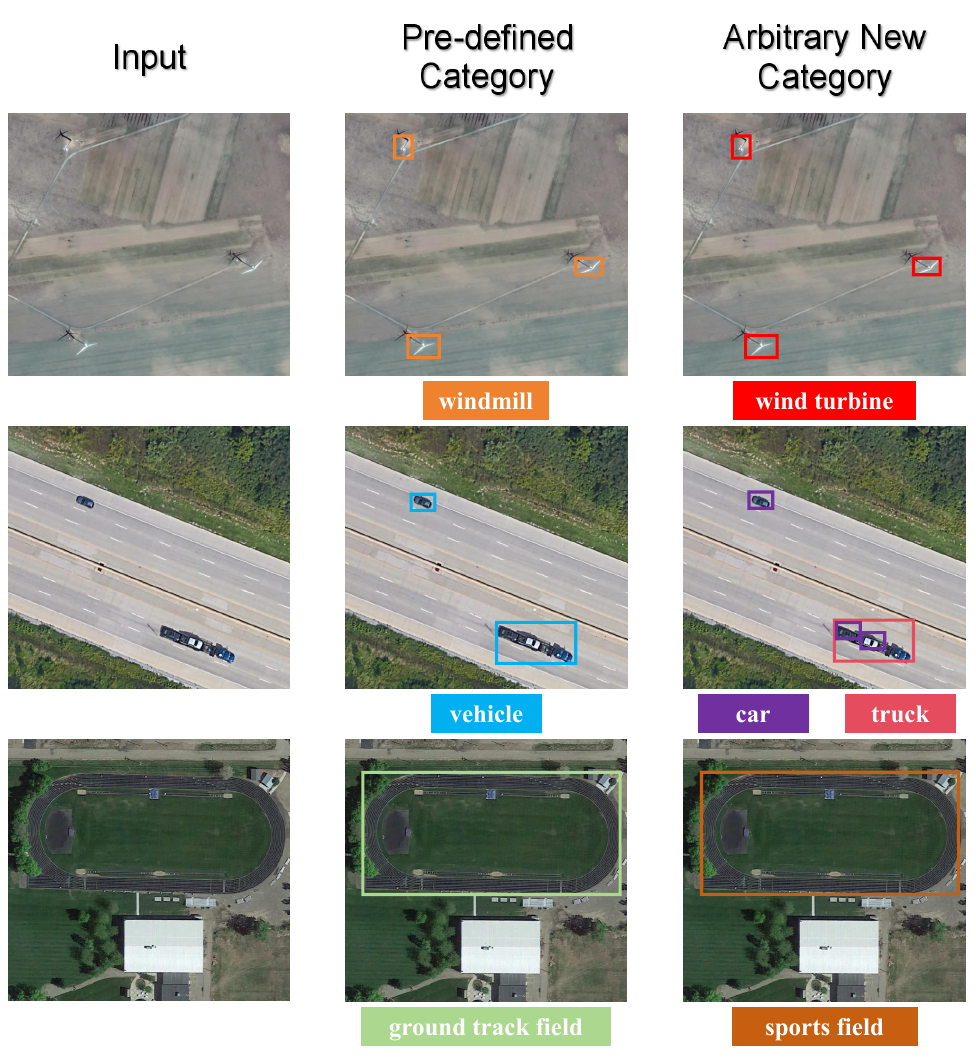}
\caption{\textbf{Comparison of traditional aerial object detection and open-vocabulary aerial object detection (OVAOD).} The former can only detect predefined categories from the training set, such as "windmill" and "vehicle," as shown in the second column. However, in practical applications, it may be necessary to detect newly defined categories, such as "wind turbine," or objects of fine-grained category descriptions, such as "car" and "truck", while OVAOD can handle this task, as shown in the third column.}
\label{fig:teaser}
\end{figure}

Aerial object detection aims to localizing objects and predicting their categories on the surface of the earth from the perspective of drones.
Despite achieving great performance in recent years, traditional aerial object detection methods still rely on pre-defined sets of training categories, and are unable to identify objects from categories absent during the training phase.

However, in real-world application scenarios, it may be necessary to detect newly designated categories, and the emergence of these new categories is often unpredictable, significantly impeding their real-world applicability (\cref{fig:teaser}).
Such a challenge has inspired the exploration of the Open-Vocabulary Aerial Object Detection (OVAOD) setting. Different from traditional closed-set aerial object detection, OVAOD methods can detect arbitrary categories described in natural language, thereby enabling the identification of previously unseen or newly defined objects without the need for retraining on fixed category sets. This flexibility makes OVAOD highly adaptable to dynamic real-world conditions, which has various practical applications such as urban management, environmental monitoring, and disaster rescue.

Due to the difficulty of acquiring diverse labeled data from the drone's viewpoint and the remarkable domain gap between aerial images and natural images such as broader range of variations in scales, orientations, and lack of  distinctive appearance features, direct application of existing open-vocabulary detection methods designed for natural scenarios is hard to achieve desired performance.
Recently, several pioneering works attempt to develop effective frameworks suitable for aerial images.
OVA-DETR~\cite{ovawei2024ova} proposes to take linguistic information as guidance and leverage the efficient detection transformer architecture RT-DETR~\cite{RTDETRzhao2024detrs} for the efficiency requirements of aerial detection.
CastDet~\cite{casdetli2024toward} presents a clip-activated teacher-student framework to learn from pseudo labels with a self-distillation and multi-teacher paradigm. 
However, previous methods tend to focus on the utilization of pretrained models for natural images and neglect to fully exploit the domain priors embedded in foundation models of more relevant domains, \textit{i.e.}, remote sensing.
Although CastDet utilizes the foundation model RemoteCLIP~\cite{remoteclipliu2024remoteclip} as semantic teacher, it only takes it as a classifier for generating pseudo category labels, failing to systematically distill the multi-level domain priors encoded in these models for more accurate detection and recognition.

In fact, remote sensing foundation models like RemoteCLIP\cite{remoteclipliu2024remoteclip} and DINOv3\cite{simeoni2025dinov3} encapsulate rich domain priors at multiple levels: RemoteCLIP's vision encoder contains \textit{visual priors} for cross-modal alignment, its text encoder embeds \textit{textual priors} about category relationships, while DINOv3 provides \textit{fine-grained spatial priors} for local feature extraction. However, existing methods only leverage surface-level classification capabilities, leaving these deep priors largely unexploited.

To this end, we propose DisDop, a unified framework that systematically distills multi-level domain priors from remote sensing foundation models into a lightweight detector for open-vocabulary aerial object detection.
Specifically, we first introduce a teacher fusion strategy to distill \textit{visual priors}. We fuse features from RemoteCLIP and DINOv3 using a self-similarity calibration mechanism, where DINOv3's spatial structure refines RemoteCLIP's semantic features. The fused teacher model is then distilled into the detector's backbone, enabling it to inherit both cross-modal alignment capability and fine-grained local discrimination ability while maintaining computational efficiency.
Second, to distill \textit{textual priors} embedded in RemoteCLIP's text encoder, we explicitly model inter-category semantic relationships by enforcing that the geometric arrangement of visual features mirrors the semantic topology in the text space. Furthermore, to address the small object challenge in aerial images, we incorporate \textit{contextual priors} by fusing global scene semantics with local instance features, mimicking how humans leverage surrounding context to identify small objects.

Our contributions can be summarized as follows:

\begin{itemize}
\item
We propose a unified multi-level domain prior distillation framework that systematically extracts and transfers visual priors, textual priors, and contextual priors from remote sensing foundation models (RemoteCLIP and DINOv3) into a lightweight detector, going beyond surface-level feature utilization.
\item
We introduce a teacher fusion strategy with self-similarity calibration that distills complementary visual priors from RemoteCLIP and DINOv3 into the detector's backbone, enabling it to simultaneously inherit cross-modal alignment and fine-grained discrimination capabilities.
\item
We propose a textual and contextual prior distillation approach that explicitly models inter-category semantic relationships from RemoteCLIP's text encoder and incorporates global scene context to enhance small object recognition in aerial images.
\item
Extensive experiments on DIOR, DOTAv2.0, and LAE-80C benchmarks demonstrate that our DisDop consistently outperforms previous state-of-the-art methods in both open-set and closed-set detection scenarios, validating the effectiveness of our multi-level domain prior distillation framework.
\end{itemize}

\section{Related Work}
\label{sec:relatedwork}

\subsection{Open Vocabulary Detection}
The open-vocabulary detection (OVD) task focuses on detecting objects based on arbitrary text queries, extending beyond predefined category sets. With the strong generalization capabilities of vision-language pretraining models such as CLIP~\cite{clipradford2021learning} and ALIGN~\cite{alignjia2021scaling}, recent OVD methods~\cite{ViLDgu2021open, Regionclipzhong2022regionclip, deticzhou2022detecting, Promptdetfeng2022promptdet, DetProdu2022learning, Corawu2023cora, OV-DETRzang2022open} incorporate these models into object detectors. For example, ViLD~\cite{ViLDgu2021open} uses knowledge distillation from pretrained VLMs to enhance OVD capabilities. RegionCLIP~\cite{Regionclipzhong2022regionclip} trains on a region-text pair dataset, strengthening its region-level understanding ability. Similarly, LAVT~\cite{yang2022lavt} introduces language-aware fusion into vision transformers for pixel-level cross-modal alignment. Detic~\cite{deticzhou2022detecting} incorporates image classification datasets to improve vocabulary coverage. Additionally, several works~\cite{Groundingdinoliu2024grounding, glipli2022grounded, Yoloworldcheng2024yolo} unify object detection with grounded pretraining, enabling detectors to perform open-vocabulary detection. Grounding-DINO~\cite{Groundingdinoliu2024grounding} combines the DINO~\cite{Dinozhang2022dino} detector with grounded pretraining and fine-grained multimodal fusion, improving open-set detection after training on large-scale datasets. Furthermore, YOLO-World~\cite{Yoloworldcheng2024yolo} employs the YOLO framework for grounded pretraining.

\begin{figure*}[!t]
    \centering
    \includegraphics[width=\textwidth]{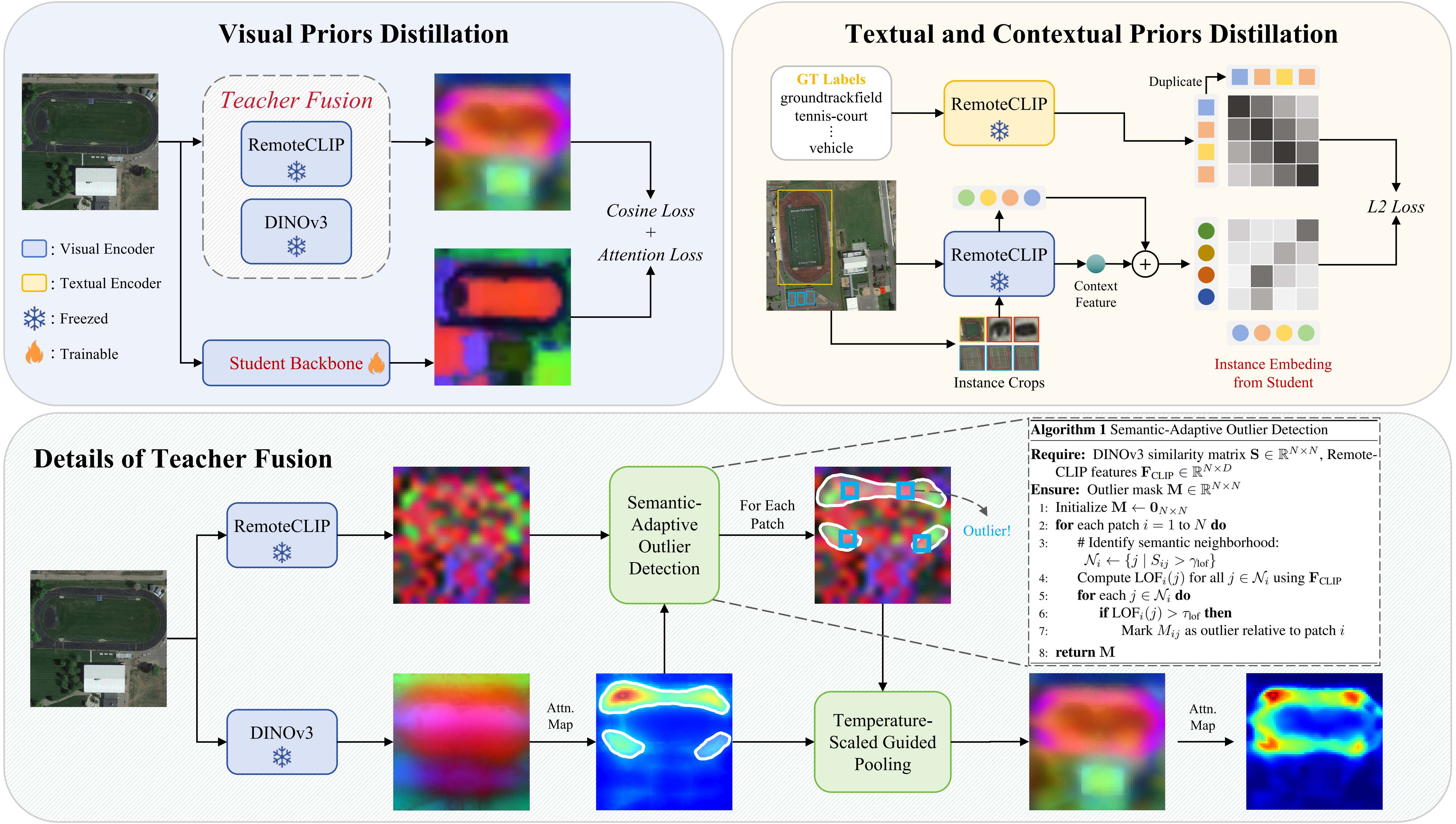}
    \caption{\textbf{Overview of DisDop Framework.} Our multi-level domain prior distillation framework consists of three key components: (1) \textit{Visual Prior Distillation}: We fuse RemoteCLIP and DINOv3 features via semantic-adaptive outlier detection and self-similarity calibration, then distill the fused knowledge into a lightweight detector backbone through cosine similarity and attention losses. (2) \textit{Textual Prior Distillation}: We distill inter-category semantic relationships from RemoteCLIP's text encoder by enforcing relational consistency between visual and text feature spaces across mini-batches. (3) \textit{Contextual Prior Distillation}: We enhance small object recognition by incorporating global scene context into the distillation targets. Through this systematic distillation pipeline, DisDop inherits both cross-modal alignment and fine-grained discrimination capabilities while maintaining computational efficiency.}
    \label{fig:pipeline}
\end{figure*}

\subsection{Aerial Object Detection}

Aerial object detection has rapidly progressed with the growth of remote sensing data and demands in fields like environmental monitoring, urban planning, and disaster response. 
Traditional category-constrained methods\cite{trad1du2023adaptive, trad2huang2022ufpmp, trad3liu2021hrdnet, trad4li2020density, trad5yang2019clustered} struggle in open-world scenarios, leading to the rise of OVAOD, which identifies novel objects without extensive annotations. 
These works highlight the importance of scalable training, domain adaptation, and vision-language integration in advancing OVAOD.
CastDet~\cite{casdetli2024toward} integrates CLIP-based knowledge distillation within a student-teacher framework to enhance pseudo-labeling.
OVA-DETR~\cite{ovawei2024ova} employs image-text alignment and transformer-based architectures for efficient and scalable detection, excelling in zero-shot mAP on DIOR.
LAE-DINO~\cite{pan2024locate} bridges domain gaps through the large-scale LAE-1M dataset and cross-modal alignment. 
\section{Method}
\label{sec:method}

\subsection{Overview}

The overall architecture of DisDop is illustrated in \cref{fig:pipeline}. We build upon the LAE detector~\cite{pan2024locate} and propose a unified \textbf{Multi-level Domain Prior Distillation} framework that systematically extracts and transfers domain priors from remote sensing foundation models (RemoteCLIP and DINOv3) into a lightweight detector.
First, we distill \textit{visual priors} from a fused teacher model into the detector's backbone (\cref{sec:backbone_distill}), enabling it to inherit both cross-modal alignment and fine-grained discrimination capabilities.
Second, we distill \textit{textual priors} from RemoteCLIP's text encoder while incorporating \textit{contextual priors} to enhance small object recognition. Through this systematic distillation of multi-level domain priors, DisDop achieves superior open-vocabulary detection performance.

\subsection{Visual Prior Distillation}
\label{sec:backbone_distill}

Open-vocabulary aerial object detection requires the detector to possess both cross-modal alignment capabilities (to connect visual features with textual descriptions) and fine-grained local feature extraction abilities (to discriminate small, densely-distributed objects). However, these two capabilities are rarely present simultaneously in a single pre-trained model. Vision-language models like RemoteCLIP, trained via contrastive learning at the image level, excel at cross-modal alignment but exhibit relatively weak local visual feature extraction due to their image-level optimization objective. Conversely, self-supervised models like DINOv3, trained via self-distillation on large-scale image data, capture fine-grained local visual features with strong semantic consistency and spatial structure, but lack the ability to align with textual descriptions.

To systematically distill these complementary \textit{visual priors} into a lightweight detector backbone, we propose a two-stage framework: (1) we first fuse features from RemoteCLIP and DINOv3 into a unified teacher representation via self-similarity calibration, where DINOv3's spatial structure refines RemoteCLIP's semantic features; (2) we then distill this fused visual prior knowledge into the detector's backbone for efficient deployment.

\noindent{\textbf{Teacher Feature Fusion via Self-Similarity Calibration.}}
Given an input image, we extract features from both pre-trained models, RemoteCLIP and DINOv3. The key innovation lies in using DINOv3's self-similarity structure to calibrate RemoteCLIP features. We first compute the spatial self-similarity matrix from DINOv3 features as $\mathbf{S} = \frac{\mathbf{F}_{\text{DINO}} \cdot \mathbf{F}_{\text{DINO}}^T}{\|\mathbf{F}_{\text{DINO}}\| \cdot \|\mathbf{F}_{\text{DINO}}^T\|} \in \mathbb{R}^{N \times N}$, where $N$ is the total number of visual patched tokens. A critical challenge in feature fusion is handling noisy or semantically inconsistent patches that can degrade the quality of calibrated features. Unlike previous methods that perform outlier detection globally across all patches using a fixed spatial or feature-based criterion, we propose a novel \textit{semantic-adaptive outlier detection} mechanism. The key insight is that whether a patch is an outlier should not be determined absolutely, but rather \textit{relative to a specific semantic context}. Concretely, for each patch $i$, we first identify its semantic neighborhood $\mathcal{N}_i = \{j | S_{ij} > \gamma_{\text{lof}}\}$ using DINOv3's self-similarity matrix, where $\gamma_{\text{lof}}$ is a similarity threshold. Within this semantically coherent region, we then apply Local Outlier Factor (LOF) analysis on RemoteCLIP features to detect patches that are visual anomalies \textit{relative to this specific semantic context}. This produces a context-dependent outlier mask $\mathbf{M} \in \mathbb{R}^{N \times N}$, where $M_{ij} = 1$ indicates that patch $j$ is an outlier with respect to patch $i$'s semantic neighborhood. Importantly, the same patch $j$ may be considered normal for one semantic context (patch $i$) but anomalous for another (patch $k$), enabling fine-grained, context-aware filtering. This semantic-adaptive strategy is fundamentally different from global outlier detection methods that treat outliers as absolute labels, and it is particularly effective for remote sensing images containing diverse objects and complex backgrounds. After filtering outliers, we apply temperature-scaled softmax to convert filtered similarities into attention weights: $\mathbf{A}_{ij} = \frac{\exp(S_{ij} / \tau)}{\sum_k \exp(S_{ik} / \tau)}$ if $S_{ij} > \gamma$ and $M_{ij} = 0$, and $0$ otherwise, where $\gamma$ filters weak correlations and $\tau$ controls attention sharpness. The final fused teacher features are obtained via weighted aggregation: $\mathbf{F}_{\text{teacher}} = \mathbf{A} \cdot \mathbf{F}_{\text{CLIP}}$. This calibration process reinforces intra-object feature consistency, where patches belonging to the same object (identified by DINOv3's self-similarity) obtain more consistent RemoteCLIP features while preserving CLIP's semantic information.

    

\noindent{\textbf{Feature Distillation to Student Backbone.}}
Although the fused model described above demonstrates excellent feature representation capabilities, its high computational complexity and the requirement to load two models make it difficult to directly apply to practical detection tasks. To address this, we treat the fused model as a teacher model and transfer its rich visual semantic information to a lightweight student model—the detector's backbone network—through knowledge distillation. Specifically, we extract corresponding intermediate layer features from both the teacher and student models, then align the student model features to the teacher model features through channel transformation and interpolation. We employ two types of knowledge distillation losses.
First is the cosine similarity loss $\mathcal{L}_{\text{cosine}} = \frac{1}{N} \sum_{i=1}^{N} \left(1 - \frac{\mathbf{f}_{\text{teacher}}^{(i)} \cdot \mathbf{f}_{\text{student}}^{(i)}}{\|\mathbf{f}_{\text{teacher}}^{(i)}\| \cdot \|\mathbf{f}_{\text{student}}^{(i)}\|}\right)$ that ensures feature direction alignment.
However, we observe that teacher model features suffer from feature homogenization, where feature vectors at different spatial locations become overly similar. Interestingly, when we apply low-temperature softmax to the similarity matrix of a patch, the resulting similarity distribution perfectly aligns with the visual semantic distribution. Therefore, we adopt an optional attention distillation loss that transfers relational structure captured by attention patterns: $\mathcal{L}_{\text{attn}} = \text{KL}\left(\mathbf{P}_{\text{teacher}} \| \mathbf{P}_{\text{student}}\right)$, where $\mathbf{P}_{\text{teacher}} = \text{softmax}(\mathbf{F}_{\text{teacher}} \mathbf{F}_{\text{teacher}}^T / \tau_t)$ and $\mathbf{P}_{\text{student}} = \text{softmax}(\mathbf{F}_{\text{student}} \mathbf{F}_{\text{student}}^T / \tau_s)$ with temperatures $\tau_t = 0.1$ and $\tau_s = 1.0$. The lower teacher temperature sharpens attention distributions to emphasize important relationships, while the higher student temperature smooths learning to prevent overfitting. The total loss combines both components as $\mathcal{L}_{\text{backbone}} = \lambda_{\text{cosine}} \mathcal{L}_{\text{cosine}} + \lambda_{\text{attn}} \mathcal{L}_{\text{attn}}$. Through this framework, we enable the detector's backbone to simultaneously learn open-vocabulary capabilities inherited from RemoteCLIP and local feature extraction abilities from DINOv3, while maintaining the computational efficiency required for practical deployment.

\begin{figure}[t]
\centering
\includegraphics[width=\linewidth]{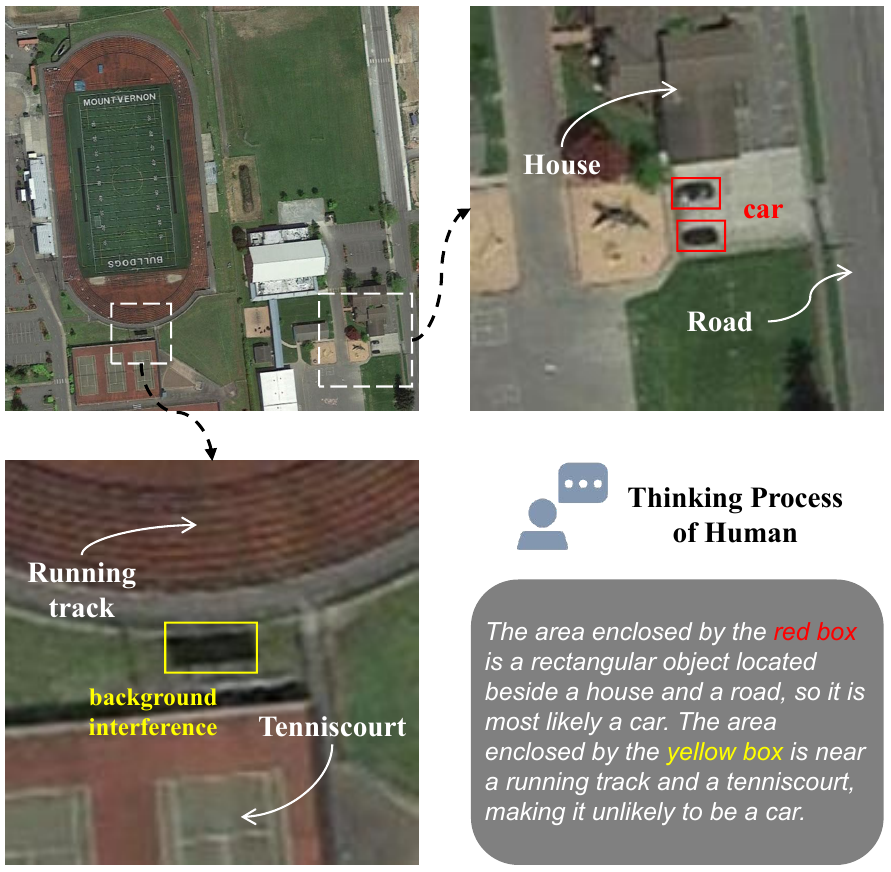}
\caption{\textbf{Motivation of global context enhancement for regional visual feature.} Humans tend to leverage contextual scene information surrounding the target region to assist in their recognition process of small objects. For example, in the upper right image, the car's small size makes it difficult to determine its category, but humans can utilize the surrounding road and houses to recognize it as a car. In the lower left image, there exists a background interference region whose local features closely resemble a car. However, humans can infer that it is not a car based on its location between a running track and a tenniscourt.}
\label{fig:CGKDMoti}
\end{figure}

\subsection{Textual and Contextual Prior Distillation}
\label{sec:textual_distill}
To enable the model to recognize objects from unseen categories during inference, it is essential to distill the \textit{textual priors} embedded in RemoteCLIP's text encoder, which encodes inter-category relationships in the language space. A common approach is to align the model's output category features with RemoteCLIP's feature space, allowing classification of new categories by computing similarities with text embeddings. While previous studies often align model features with discrete text embeddings, this can lead to overfitting and neglect the rich diversity of intra-class visual features. A more robust baseline aligns the model's output category features ($f_c^i$) directly with the RemoteCLIP visual features ($f_v^i$) extracted from the corresponding region of interest (ROI):
\begin{equation}
L_{KD} = \frac{1}{N} \sum_{i=1}^{N} \left( f_c^i - f_v^i \right)^2 \label{eq:kd1}
\end{equation}
Here, $N$ is the total number of ROIs in a training batch. By minimizing this loss, we encourage the model to learn a continuous and generalizable visual feature space.

However, this instance-level visual feature distillation, while an improvement, has two significant limitations. First, it struggles with small objects. Cropped regions from extremely small bounding boxes (e.g., 2×2 pixels) often lack sufficient detail, leading to poor-quality visual features ($f_v^i$) from RemoteCLIP that deviate from its original training distribution. This is contrary to human cognition, where we leverage surrounding scene context to identify small objects (\cref{fig:CGKDMoti}). For example, a tiny, ambiguous shape on a road next to a house is likely inferred to be a "car" due to its context, not just its local features. Second, and more critically for distilling \textit{textual priors}, this point-wise alignment overlooks the relative semantic relationships between different categories encoded in RemoteCLIP's text encoder. The loss in \cref{eq:kd1} only forces each instance's feature to match its visual target independently, failing to model the semantic topology across categories. This can lead to feature space inconsistencies, where the learned distance between features of semantically related categories (e.g., “bridge” and “ship”) might be larger than that of unrelated categories (e.g., “bridge” and “windmill”), degrading open-vocabulary classification performance which relies on semantic similarity.

To address both limitations and effectively distill \textit{textual priors} from RemoteCLIP's text encoder while incorporating \textit{contextual priors} for small object recognition, we enhance the distillation target by incorporating global scene context and explicitly model the semantic relationships between categories encoded in the text space. Our distillation loss is formulated as:
\begin{equation}
L_{distill} = \frac{1}{N^2} \sum_{i=1}^{N} \sum_{j=1}^{N} \left( \left\| f_c^i - f_{eh}^j \right\|_{cos} - \left\| t^i - t^j \right\|_{cos} \right)^2
\end{equation}
This formulation introduces two key designs that enable multi-level prior distillation. First, to distill \textit{contextual priors} and tackle the small-object problem, we create an enhanced visual feature ($f_{eh}^j$) that fuses instance-level features with global context:
\begin{equation}
f_{eh}^j = LN\left( \mu f_g^j + (1 - \mu) f_v^j \right)
\end{equation}
Here, $f_v^j$ is the local visual feature from the ROI crop, $f_g^j$ is the global visual feature extracted from the entire image containing the ROI, and $\mu \in [0, 1]$ is a learnable parameter that dynamically balances their contribution. This ensures the distillation target contains both fine-grained details and scene-level contextual semantics.

Second, to distill \textit{textual priors} from RemoteCLIP's text encoder, the loss function enforces a relational consistency constraint. It requires that the cosine distance between the model's output feature for instance $i$ ($f_c^i$) and the enhanced visual feature for instance $j$ ($f_{eh}^j$) mirrors the cosine distance between their corresponding category text embeddings ($t^i$ and $t^j$) in RemoteCLIP's language space. This forces the model to learn a feature space where the geometric arrangement of visual features aligns with the semantic topology encoded in the text encoder. For example, the features for "bridge" and "ship" are encouraged to be closer to each other than the features for "bridge" and "windmill", preserving the semantic structure of the language space.

Furthermore, inspired by \cite{gkchan2023global}, but unlike their approach which operates only within a single image, we perform this relational distillation across the entire mini-batch. Since most training images contain few object categories, within-image distillation provides sparse supervision for modeling semantic relationships. By leveraging the diversity of categories across a mini-batch, our approach receives a much denser and more effective supervision for distilling textual priors, significantly enhancing its ability to transfer the inter-category semantic topology from RemoteCLIP's text encoder and improve generalization to unseen categories.

\subsection{Training}

Our training pipeline consists of two sequential stages that systematically distill multi-level domain priors into the detector. In the first stage, we train only the backbone network only using the images from LAE-1M dataset~\cite{pan2024locate}, distilling fused visual knowledge from RemoteCLIP and DINOv3 into a lightweight Swin Transformer. In the second stage, we train the full detector on the labeled LAE-1M dataset, incorporating textual and contextual prior distillation. This two-stage approach enables efficient knowledge transfer while maintaining computational feasibility, as the heavy teacher models (RemoteCLIP and DINOv3) are only needed during the first stage.

\noindent\textbf{Stage I: Visual Prior Distillation.}
In the first stage, we train the detector's backbone network on the unlabeled LAE-1M dataset using the combined loss $\mathcal{L}_{\text{backbone}} = \lambda_{\text{cosine}} \mathcal{L}_{\text{cosine}} + \lambda_{\text{attn}} \mathcal{L}_{\text{attn}}$ described in \cref{sec:backbone_distill}. The fused teacher model is frozen while only the student backbone (Swin Transformer Tiny) is updated.

\noindent\textbf{Stage II: Detector Training with Textual and Contextual Prior Distillation.}
After obtaining the distilled backbone, we train the full detector on the labeled LAE-1M dataset.
The training objective combines standard detection losses with the textual and contextual prior distillation loss $\mathcal{L}_{\text{distill}}$ described in \cref{sec:textual_distill}.
\section{Experiments}
\definecolor{mygray}{gray}{.9}

\subsection{Experimental Setup}

\begin{table*}[t]
\caption{\textbf{The open-set detection results on DIOR, DOTAv2.0 and LAE-80C benchmarks.}}
\centering
\resizebox{0.73\linewidth}{!}{%
\begin{tabular}{l|c|c|c|c}
\toprule
\multirow{2}{*}{\textbf{Method}} & \multirow{2}{*}{\textbf{Pre-Training Data}} & \multicolumn{1}{c|}{\textbf{DIOR}} & \multicolumn{1}{c|}{\textbf{DOTAv2.0}} & \multicolumn{1}{c}{\textbf{LAE-80C}} \\
 &  & $AP_{50}$ & $mAP$  & $mAP$ \\ \hline
\multicolumn{1}{l|}{GLIP with \textit{DVC} \cite{li2022grounded}} & LAE-1M & 82.8  & 43.0  & 16.5   \\
\multicolumn{1}{l|}{GroundingDINO with \textit{DVC} \cite{li2022grounded}} & LAE-1M & 83.6  & 46.0 & 17.7  \\ 
\multicolumn{1}{l|}{LAE-DINO \cite{pan2024locate}} & LAE-1M & 85.5 & 46.8  & 20.2   \\
\rowcolor{mygray}\multicolumn{1}{l|}{\textbf{DisDop} (Ours)} & LAE-1M & \textbf{87.2} & \textbf{47.5}  & \textbf{22.5}   \\

\bottomrule
\end{tabular}}

\label{tab:table2}
\end{table*}

\begin{table*}[t]
\caption{\textbf{The closed-set detection results on on DIOR and DOTAv2.0 test set.}}
\centering
\resizebox{0.8\linewidth}{!}{%
\begin{tabular}{lcccc}
\toprule
\multicolumn{1}{l|}{\multirow{2}{*}{\textbf{Method}}} & \multicolumn{1}{c|}{\multirow{2}{*}{\textbf{Backbone}}} & \multicolumn{1}{c|}{\multirow{2}{*}{\textbf{Pre-Training Data}}} & \multicolumn{2}{c}{\textbf{Fine-Tuning}} \\
\multicolumn{1}{l|}{} & \multicolumn{1}{c|}{} & \multicolumn{1}{c|}{} & DIOR($AP_{50}$) & DOTAv2.0($mAP$)  \\ \hline
\multicolumn{5}{l}{\textit{Generic   Object Detection}} \\ \hline
\multicolumn{1}{l|}{GASSL \cite{ayush2021geography}} & \multicolumn{1}{c|}{ResNet-50} & \multicolumn{1}{c|}{-} & 67.4 & - \\
\multicolumn{1}{l|}{CACO \cite{mall2023change}} & \multicolumn{1}{c|}{ResNet-50} & \multicolumn{1}{c|}{Sentinel-2} & 66.9 & - \\
\multicolumn{1}{l|}{TOV \cite{tao2023tov}} & \multicolumn{1}{c|}{ResNet-50} & \multicolumn{1}{c|}{TOV-NI,TOV-R} & 70.2 & - \\
\multicolumn{1}{l|}{Scale-MAE \cite{reed2023scale}} & \multicolumn{1}{c|}{ViT-L} & \multicolumn{1}{c|}{FMoW} & 73.8 & - \\
\multicolumn{1}{l|}{SatLas \cite{bastani2023satlaspretrain}} & \multicolumn{1}{c|}{Swin-B} & \multicolumn{1}{c|}{ SatlasPretrain } & 74.1 & -  \\
\multicolumn{1}{l|}{RingMo \cite{sun2022ringmo}} & \multicolumn{1}{c|}{Swin-B} & \multicolumn{1}{c|}{RingMoPretrain} & 75.9 & -  \\
\multicolumn{1}{l|}{SkySense \cite{guo2024skysense}} & \multicolumn{1}{c|}{Swin-H} & \multicolumn{1}{c|}{multi-modal RSI} & 78.7 & -  \\
\multicolumn{1}{l|}{MTP \cite{wang2024mtp}} & \multicolumn{1}{c|}{Swin-H} & \multicolumn{1}{c|}{MillionAID} & 81.1 & -  \\
\hline
\multicolumn{5}{l}{\textit{Open-Vocabulary   Object Detection}} \\ \hline
\multicolumn{1}{l|}{GLIP-FT \cite{li2022grounded}} & \multicolumn{1}{c|}{Swin-T} & \multicolumn{1}{c|}{LAE-1M} & 88.9 & 51.5  \\
\multicolumn{1}{l|}{GroundingDINO-FT \cite{liu2024grounding}} & \multicolumn{1}{c|}{Swin-T} & \multicolumn{1}{c|}{LAE-1M} & 91.1 & 55.1 \\
\multicolumn{1}{l|}{LAE-DINO-FT\cite{liu2024grounding}} & \multicolumn{1}{c|}{Swin-T} & \multicolumn{1}{c|}{LAE-1M} & 92.2 & 57.9  \\
\rowcolor{mygray}\multicolumn{1}{l|}{\textbf{DisDop-FT} (Ours)} & \multicolumn{1}{c|}{Swin-T} & \multicolumn{1}{c|}{LAE-1M} & \textbf{92.6} & \textbf{58.6}  \\

\bottomrule
\end{tabular}}
 \label{tab:table3}
\end{table*}

\subsubsection{Evaluation}
To validate the effectiveness of our proposed DisDop model, we employ DIOR \cite{diorli2020object} and DOTAv2.0 \cite{Xia_2018_CVPR}, two widely adopted benchmarks in the remote sensing community. 
Note that we only evaluate the horizontal bounding box detection performance on DOTAv2.0. Additionally, we evaluate on LAE-80C~\cite{pan2024locate}, a remote sensing OVD benchmark comprising 80 classes, to thoroughly assess the open-set detection capability. These three benchmarks collectively enable comprehensive evaluation of both open-set and closed-set detection performance. We adopt $mAP$ and $AP_{50}$ as evaluation metrics throughout our experiments.

\subsubsection{Implementation Details.}

We implement our framework using MMDetection~\cite{chen2019mmdetection} and PyTorch. For the teacher model, we use DINOv3-ViT-L/16~\cite{simeoni2025dinov3} and RemoteCLIP-ViT-L/14~\cite{remoteclipliu2024remoteclip} as the foundation models. The student backbone is a random initialized Swin Transformer-Tiny~\cite{liu2021swin}. In Stage I, we train for 15,000 iterations with batch size 8 using the AdamW optimizer. For the semantic-adaptive outlier detection, we set the similarity thresholds $\gamma = 0.5$ and $\gamma_{\text{lof}} = 0.8$, LOF threshold $\tau_{\text{lof}} = 1.2$. In Stage II, we follow the training configuration of LAE~\cite{pan2024locate} with the detection-specific hyperparameters. All experiments are conducted on 8 NVIDIA A100 GPUs.

\subsection{Detection Results}
\subsubsection{Open-Set Detection.}
We compare the open-set detection results with three effective OVD methods, GLIP \cite{li2022grounded}, GroundingDINO \cite{liu2024grounding} and LAE-DINO\cite{pan2024locate}, trained on natural and remote sensing scenes datasets as shown in Table \ref{tab:table2}. 
To train on LAE-1M dataset, we introduce DVC, which was proposed in LAE-DINO, to GLIP and GroundingDINO.  

As shown in Table \ref{tab:table2}, our DisDop model achieves superior performance across all three benchmarks. On DIOR, DisDop achieves 87.2\% $AP_{50}$, outperforming LAE-DINO by 1.7\% and GroundingDINO with DVC by 3.6\%. For DOTAv2.0, our method reaches 47.5\% mAP, demonstrating a 0.7\% improvement over LAE-DINO and a 1.5\% gain over GroundingDINO with DVC. Most notably, on the LAE-80C benchmark which evaluates detection capability across 80 diverse categories, DisDop achieves 22.5\% mAP, significantly surpassing LAE-DINO by 2.3\% and GroundingDINO by 4.8\%. These consistent improvements across different benchmarks demonstrate the effectiveness of our distillation-based approach in enhancing open-vocabulary detection capabilities for remote sensing imagery. The larger performance gains on LAE-80C particularly highlight DisDop's superior generalization ability to novel and diverse object categories.

\subsubsection{Closed-Set Detection.}
To prove the benefits of OVD, we perform fine-tuning experiments in remote sensing scenes, comparing some generic detectors (GD) on DIOR and DOTAv2.0 datasets as shown in Table \ref{tab:table3}. Most previous GDs are fine-tuned to object detection datasets after pre-training on remote-sensing images using a self-supervised approach. We directly cite the original paper results due to the lack of open source for these generic detectors.

Table \ref{tab:table3} demonstrates that OVD methods significantly outperform traditional generic detectors when fine-tuned on downstream tasks. Our DisDop achieves 92.6\% $AP_{50}$ on DIOR and 58.6\% mAP on DOTAv2.0, surpassing the previous best OVD method LAE-DINO-FT by 0.4\% and 0.7\% respectively. Compared to generic detectors, DisDop shows substantial advantages: it outperforms MTP (the best generic detector) by 11.5\% on DIOR, despite MTP using a much larger Swin-H backbone. This significant performance gap highlights the superiority of OVD pre-training over self-supervised pre-training approaches. The superior performance stems from OVD's ability to learn rich semantic representations through vision-language alignment during pre-training, which provides a stronger foundation for downstream task adaptation. Furthermore, our distillation-based approach effectively transfers knowledge from the teacher model while maintaining computational efficiency, enabling better feature learning and cross-modal understanding for remote sensing object detection. 

\subsection{Ablation Studies}
\subsubsection{Component Analysis.} We conduct ablation experiments on DIOR test set to validate the effectiveness of each proposed component as shown in Table \ref{tab:ablation_compo}. The baseline model is LAE-DINO \cite{pan2024locate} without any distillation strategies.
\begin{table}[h]
\centering
\caption{\textbf{Ablation study Results}. The best results for each metric are \textbf{bolded}. VPD denote the Visual Prior Distillation. TCPD denote the Textual and Context Prior Distillation.}
\begin{tabular}{c c|c c c}
\toprule
\multirow{2}{*}{VPD} & \multirow{2}{*}{TCPD} & \multicolumn{1}{c}{\textbf{DIOR}} & \multicolumn{1}{c}{\textbf{DOTAv2.0}} & \multicolumn{1}{c}{\textbf{LAE-80C}} \\
 &  & $AP_{50}$ & $mAP$ & $mAP$ \\
\hline
 &  & 85.5 & 46.8 & 20.2 \\
$\checkmark$ &  & 86.6 & 47.2 & 21.5 \\
 & $\checkmark$ & 86.2 & 47.1 & 20.9 \\
    $\checkmark$ & $\checkmark$ & \textbf{87.2} & \textbf{47.5} & \textbf{22.5} \\
\bottomrule
\end{tabular}
\label{tab:ablation_compo}
\end{table}

\subsubsection{Ablation on Teacher Feature Fusion Strategy.}
To validate the effectiveness of our semantic-adaptive outlier detection mechanism in the teacher feature fusion process, we compare it with alternative fusion strategies in Table \ref{tab:ablation_fusion}. We compare with a global outlier detection method that applies LOF across all patches uniformly, and a simple attention-based fusion using only similarity thresholding without outlier filtering.

\begin{table}[h]
\centering
\caption{\textbf{Ablation study on teacher feature fusion strategies.} SAOD denotes Semantic-Adaptive Outlier Detection.}
\begin{tabular}{l|c c c}
\toprule
\multirow{2}{*}{Fusion Strategy} & \multicolumn{1}{c}{\textbf{DIOR}} & \multicolumn{1}{c}{\textbf{DOTAv2.0}} & \multicolumn{1}{c}{\textbf{LAE-80C}} \\
 & $AP_{50}$ & $mAP$ & $mAP$ \\
\hline
w/o Outlier Filtering & 86.4 & 47.1 & 21.3 \\
Global LOF & 86.3 & 47.0 & 21.2 \\
SAOD (Ours) & \textbf{86.6} & \textbf{47.2} & \textbf{21.5} \\
\bottomrule
\end{tabular}
\label{tab:ablation_fusion}
\end{table}

As shown in Table \ref{tab:ablation_fusion}, our semantic-adaptive outlier detection achieves the best performance across all metrics. The global LOF method (86.3\% $AP_{50}$) performs worse than our approach because it treats outliers as absolute labels, failing to consider that a patch may be normal in one semantic context but anomalous in another. The attention fusion without outlier filtering (86.4\%) shows better results but still falls short, as noisy patches can corrupt the calibrated features. Our semantic-adaptive strategy achieves 86.6\% $AP_{50}$ by detecting outliers relative to specific semantic contexts, enabling fine-grained, context-aware filtering that is particularly effective for complex aerial scenes.

\subsubsection{Ablation on Relational Distillation Scope.}
To validate the importance of performing relational distillation across mini-batches rather than within single images, we compare different distillation scopes in Table \ref{tab:ablation_scope}. The point-wise baseline uses Equation \ref{eq:kd1} without any relational constraints. Within-image relational distillation computes the loss only among ROIs in the same image, while within-batch approach leverages the entire mini-batch diversity.

\begin{table}[h]
\centering
\caption{\textbf{Ablation study on relational distillation scope.}}
\begin{tabular}{l|c c c}
\toprule
\multirow{2}{*}{Distillation Scope} & \multicolumn{1}{c}{\textbf{DIOR}} & \multicolumn{1}{c}{\textbf{DOTAv2.0}} & \multicolumn{1}{c}{\textbf{LAE-80C}} \\
 & $AP_{50}$ & $mAP$ & $mAP$ \\
\hline
Point-Wise & 85.8 & 46.8 & 20.4 \\
Within-Image & 86.1 & 46.9 & 20.7 \\
Within-Batch (Ours) & \textbf{86.2} & \textbf{47.1} & \textbf{20.9} \\
\bottomrule
\end{tabular}
\label{tab:ablation_scope}
\end{table}

Table \ref{tab:ablation_scope} demonstrates the superiority of our cross-batch relational distillation. The point-wise baseline achieves only 85.8\% $AP_{50}$, failing to capture inter-category semantic relationships. Within-image relational distillation improves to 86.1\%, but provides sparse supervision since most images contain only a few object categories. Our cross-batch approach achieves 86.2\% $AP_{50}$ by leveraging category diversity across the entire mini-batch, providing much denser supervision for modeling the semantic topology from RemoteCLIP's text encoder. The 0.5\% improvement on LAE-80C particularly highlights the benefit for generalizing to unseen categories, validating our design choice to distill textual priors at the batch level.

\section{Conclusion}
\label{sec:conclusion}

In this paper, we presented DisDop, a unified framework for open-vocabulary aerial object detection that systematically distills multi-level domain priors from remote sensing foundation models. By introducing a teacher fusion strategy with self-similarity calibration, we effectively distill visual priors from RemoteCLIP and DINOv3 into a lightweight detector backbone. Furthermore, our textual and contextual prior distillation approach explicitly models inter-category semantic relationships and incorporates global scene context to enhance small object recognition. Extensive experiments on DIOR, DOTAv2.0, and LAE-80C benchmarks demonstrate that DisDop consistently outperforms previous state-of-the-art methods in both open-set and closed-set detection scenarios. Our work provides valuable insights into leveraging domain-specific foundation models for aerial object detection and opens new directions for knowledge distillation in aerial applications. Looking forward, we envision DisDop serving as a strong foundation for future research in efficient and effective open-vocabulary aerial detection tasks.

{
    \small
    \bibliographystyle{ieeenat_fullname}
    \bibliography{main}
}


\end{document}